\pdfoutput=1

\documentclass[11pt]{article}

\usepackage[final]{acl}

\usepackage{times}
\usepackage{latexsym}
\usepackage{tabularx}
\usepackage{booktabs}
\usepackage{multirow}
\usepackage{graphicx}
\usepackage{algorithm}
\usepackage{algpseudocode}

\usepackage{dsfont}
\usepackage{amsfonts}
\usepackage{amssymb}
\usepackage{bbm}
\usepackage{accsupp}
\usepackage[T1]{fontenc}
\usepackage[utf8]{inputenc}
\usepackage{microtype}

\usepackage{inconsolata}
\usepackage{xcolor}

\def\ours{\textit{PAC}{}}

\title{Data Contamination Calibration for Black-box LLMs}

\author{Wentao Ye\textsuperscript{1}, Jiaqi Hu\textsuperscript{1}, 
   Liyao Li\textsuperscript{1}, Haobo Wang\textsuperscript{1}, Gang Chen\textsuperscript{1},
   Junbo Zhao\textsuperscript{1}\\
   \textsuperscript{1}Zhejiang University\\
   \textit{Correspondence:} \texttt{j.zhao@zju.edu.cn}
   }

\begin{document}
\maketitle
\begin{abstract}
The rapid advancements of Large Language Models (LLMs) are tightly associated with the expansion of the training data size. 
However, the unchecked ultra-large-scale training sets introduce a series of potential risks like data contamination, i.e. the benchmark data is used for training. 
In this work, we propose a holistic method named \textbf{\underline{P}}olarized \textbf{\underline{A}}ugment \textbf{\underline{C}}alibration (\ours{}) along with a brand-new dataset named StackMIA to help detect the contaminated data and diminish the contamination effect.
\ours{} extends the popular MIA (Membership Inference Attack) --- from the machine learning community --- by forming a more global target for detecting training data to clarify invisible training data.
As a pioneering work, \ours{} is very much plug-and-play that can be integrated with most (if not all) current white- and black-box (for the first time) LLMs.
By extensive experiments, \ours{} outperforms existing methods by at least \textbf{4.5\%}, in data contamination detection on more than \textbf{4} dataset formats, with more than \textbf{10} base LLMs. Besides, our application in real-world scenarios highlights the prominent presence of contamination and related issues. \footnote{Our code is available: \href{https://github.com/yyy01/PAC}{https://github.com/yyy01/PAC}.}
\end{abstract}

\section{Introduction}

As is widely acknowledged, the rapid advancements of Large Language Models (LLMs) in natural language tasks are largely attributed to the incredible expansion of the size of the training data \cite{kaplan2020scaling}. 
Despite the massive successes, this unmanaged expansion has introduced a series of significant issues that are yet explored, particularly \emph{data contamination}.
This issue arises notably when the benchmarking data is inadvertently included in the training sets. This contamination leads to misleading evaluation results \cite{zhou2023don, narayanan2023gpt}, thus deducing difficulties in acquiring effective and secure models. 
Additionally, training on datasets with copyrighted, private, or harmful content could violate laws, infringe on privacy, and introduce biases \cite{carlini2019secret, nasr2018machine}.
Unlike in earlier stages of machine learning, we posit that this problem would be much more prevalent in the age of the LLMs, very much due to the inevitable lack of scrutinization of the much-scaled --- and often private --- training data~\cite{magar2022data, dodge2021documenting}. 

\begin{figure}[htb]
    \centering
    \includegraphics[width=\linewidth]{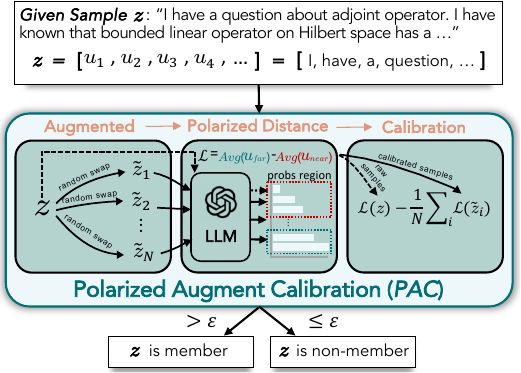}
    \caption{We focus on determining whether a given sample is contained in the training set of the LLMs. For a candidate $z$, \ours{} utilizes random swap augmentation to generate adjacent samples in local distribution regions. Consequently, \ours{} compares the polarized distance of $z$ with its adjacent samples $\tilde{z}$, where the polarized distance is a spatial measurement jointly considering far and near probability regions.} 
    \label{fig:framework}
\end{figure}

\begin{figure}[htb]
    \centering
    \includegraphics[width=\linewidth]{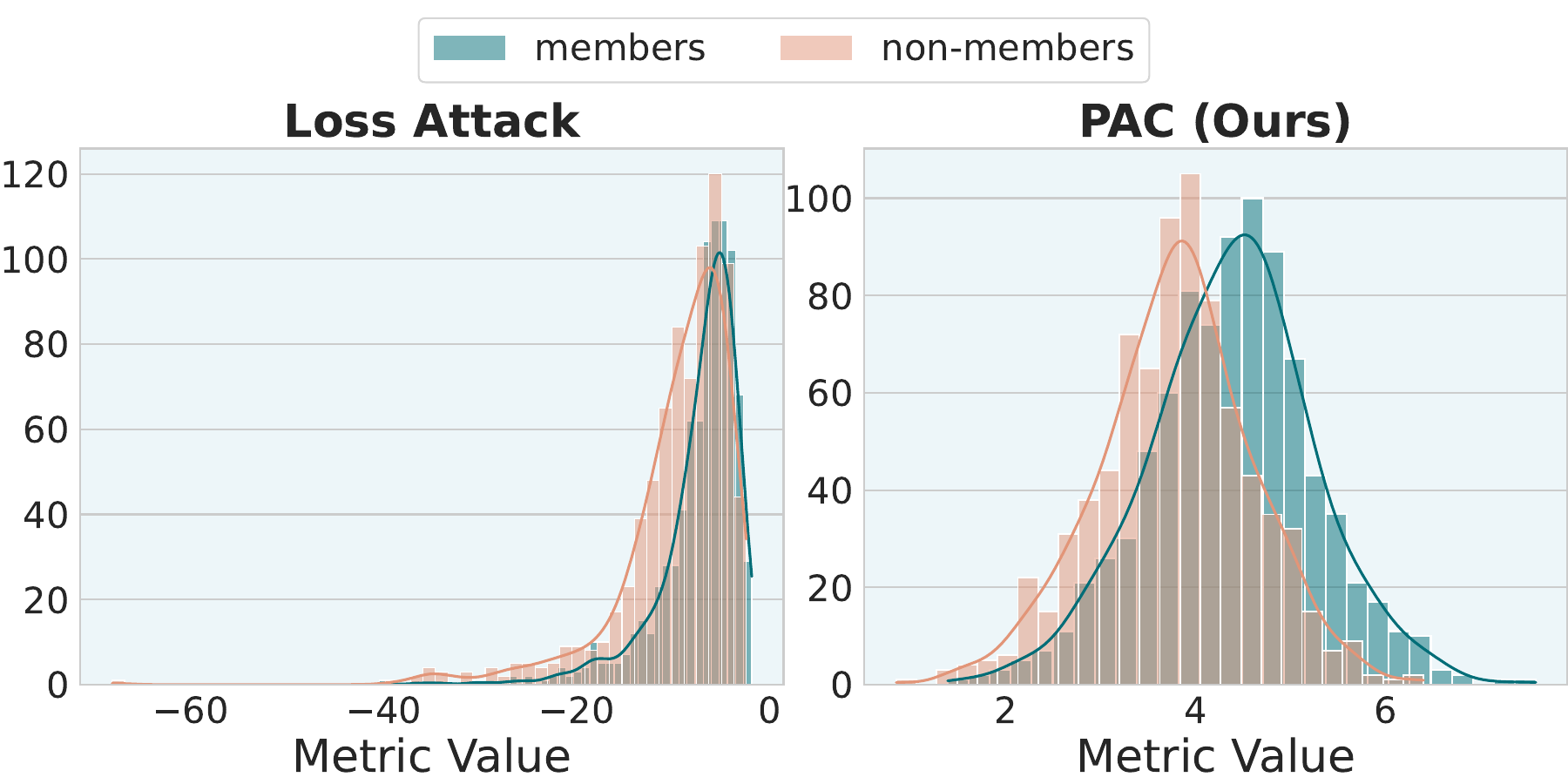}
    \caption{Histogram of the model confidence (follow the loss attack to use perplexity) before and after \ours{} in gpt-3 (davinci-002) on WikiMIA dataset \cite{shi2023detecting}, where \ours{} significantly enhances the salience of differences between members and non-members.} 
    \label{fig:intro_prob}
\end{figure}

In this work, we position the training data detection for LLMs as an extension of the \emph{membership inference attacks} (MIA) \cite{shokri2017membership} in the literature of machine learning.
MIA targets distinguishing whether the given data samples are \emph{members} (training data) or \emph{non-members} (not be trained).
The previous line of work can generally be categorized into score-based (i.e., calibration-free) \cite{yeom2018privacy, salem2019ml, shi2023detecting} and calibration-based \cite{watson2021importance, carlini2022membership, mattern2023membership} methodologies.
Despite the promise, these approaches hardly suffice for current LLMs.
The conventional setup in machine learning may generally focus on a small-scale training set, accompanied by global confidence distribution differences between members and non-members. 
However, this assumption no longer holds in LLMs, leading to a situation where non-members can also exhibit misleadingly high confidence levels, as shown in Figure~\ref{fig:intro_prob}. 
In addition, the MIA approaches generally rely on training an external reference or proxy model using member data distribution approximated to the targeted model. This, unfortunately, has contradicted the black-box setup of many current LLMs.

With this paper, we make two major efforts: (i)-a holistic scheme named polarized augment calibration (\ours{}) to resolve the challenges of black-box training data detection and (ii)-a brand new dataset for newly released LLMs.

Notably, the current related methods mostly assess the overfitting metric at individual points. Upon revisiting the issue of typical MIA, we find that the geometric properties of the samples may better reflect the latent differences introduced by training. For members, they should exhibit two characteristics: (i) high confidence at individual points; and (ii) being part of a poorly generalized local manifold. In other words, if a sample lies in a poor-generalization region but shows anomalous confidence, it suggests that the model is merely memorizing (or overfitting) rather than truly generalizing.

Based on such observations, as depicted in Figure~\ref{fig:framework}, we propose \textbf{P}olarized \textbf{A}ugment \textbf{C}alibration (\ours{}), which explores the confidence discrepancies in local regions through data augmentation techniques, aimed at detecting those overfitted training samples.
Upon this, to address the bias of existing global confidence metrics, we have developed a brand-new evaluation score named \emph{polarized distance}, as a polarized calculation of \emph{flagged tokens} with far and near local regions in the probability space.
We further provide an in-depth explanation of \ours{} detection from a theoretical perspective.
Compared to previous relevant methods relying on fully probability-accessed LLMs, we introduce a new probabilistic tracking method to extract probabilities under limited conditions(e.g., OpenAI API) for the first time.2
This becomes the trailblazers to extend our detection to a black-box setup where only access to partial probabilities is permitted, enabling \ours{} to be applied to almost all LLMs. 

For the benchmark, the current available benchmark for the contamination problem is very limited. We construct and introduce StackMIA tailored for reliable and scalable detection of the latest LLMs. It ensures reliability by adopting a time-based member/non-member classification similar to \cite{shi2023detecting}, where members are visible during pre-training and non-members are not.
StackMIA is dynamically updated, by offering detailed timestamps in order to quickly adapt to any up-to-date LLM through our curation pipeline.
By contrast, the existing benchmark WikiMIA\cite{shi2023detecting} does not possess these properties, making it hard to use, especially on the LLMs released post-2023.

Last but not least, we exhibit extensive experiments on 10 commonly used models to evaluate the \ours{} against six existing major baselines. 
These results demonstrate that \ours{} outperforms the strongest existing baseline by \textbf{5.9\%} and\textbf{ 4.5\%} in the AUC score respectively on StackMIA and WikiMIA. 
\ours{} further expresses superior robustness under conditions of ambiguous memory or detection of fine-tuning data. 
To further validate the effectiveness of \ours{} in real-world applications, we provide a set of case studies for data contamination and different security risks \cite{carlini2019secret, nasr2018machine, nasr2018comprehensive} on ChatGPT and GPT-4. 
The conclusions from these examples highlight the ubiquity of security risks upon widespread deployment \cite{ye2023assessing}. 

Our contributions can be summarized as: 
\begin{enumerate}
    \item We introduce \ours{}, an innovative, theory-supported MIA method for black-box LLMs that does not rely on external models, consistently outperforming leading approaches.
    \item Our black-box probability extraction algorithm makes \ours{} a trailblazer for LLMs with restricted probability access. We also unveil the StackMIA benchmark, addressing the gaps in MIA datasets for pretraining phrases.
    \item Applying \ours{} to ChatGPT and GPT-4 highlights the prevalent data contamination issue, prompting a call to the academic community for solutions to ensure safer, more dependable LLMs.
\end{enumerate}

\section{Related Work}

\textbf{Data Contamination.} 
As \citet{magar2022data} mentioned, data contamination is the infiltration of downstream test data into the pretraining corpus, which may seriously mislead evaluation results. \citet{dodge2021documenting} and \citet{brown2020language} highlighted the tangible presence of contamination issues in models (e.g. GPT-3) and corpora (e.g. C4). Such contamination risks models memorizing rather than learning, affecting their exploitation \citep{magar2022data}. Detection methods have been proposed, using n-gram overlap ratios against pre-training data to identify contaminated samples \citep{du2022glam,wei2021finetuned,chowdhery2023palm}. 
However, these existing methods rely on access to the pretraining corpus, which is unfeasible for many LLMs. 
Recent studies \cite{golchin2023time, weller2023according} shifted towards a more common black-box setting, by extracting outputs with trigger tokens \cite{carlini2021extracting} or specified prompts \cite{sainz2023did, nasrscalable} from LLMs as contaminated samples and comparing with the test set.
Unfortunately, the extracted samples are usually too broad, making these methods ineffective for detecting given samples.

\textbf{Membership Inference Attack.} 
\label{section:rw_mia}
Membership Inference Attacks (MIA), proposed by \citet{shokri2017membership}, is defined as determining whether a given sample is part of the training set. MIA is predicated on the models' inevitable overfitting \cite{yeom2018privacy}, leading to a differential performance on training samples (members) versus non-members. MIA has been applied in privacy protection \cite{jayaraman2019evaluating, zanella2020analyzing, nasr2021adversary, nasr2023tight, steinke2023privacy}, machine-generated text detection \cite{mitchell2023detectgpt, solaiman2019release}, DNA inference \cite{zerhouni2008protecting}, etc. Given the limited access to most current LLMs, research has concentrated on black-box conditions. Typical score-based methods employ loss or partial token probabilities \cite{shi2023detecting}, though these may incorrectly flag non-members \cite{carlini2022membership}. \citet{watson2021importance} proposed calibration-based methods to utilize a difficulty calibration score to regularize raw scores. Specifically, \citet{carlini2021extracting}, \citet{ye2022enhanced}, and \citet{mireshghallah2022quantifying} train reference models to rectify anomalies with the average of different models. 
\citet{mattern2023membership} might be most closely aligned with our work, utilizing additional models to generate similar samples for calibration purposes. while Yet, the practicality is constrained by their need for extra models. Furthermore, obtaining the necessary probabilities for MIA is challenging in many recent LLMs.
To our knowledge, we are the pioneers in developing a calibration-based detection method jointly considering the calibration of global confidence and local distribution.

\section{Problem Definition and Preliminary Work}

We first provide a comprehensive formal definition of the training data detection for LLMs in the context of a two-stage training process \cite{kenton2019bert}. Upon this foundation, we introduce a new dynamic data benchmark (Section ~\ref{subsection:benchmark}) that can be applied to recently released LLMs. By open-sourcing the benchmark upon publication, we expect to use it to foster further research in the community.

\subsection{Problem Formulation}
\label{subsection:formulation}

\textbf{Two-stage training.} The training process of LLMs consists of two stages: unsupervised pre-training on a large-scale corpora, 
and supervised fine-tuning of labeled data for downstream tasks.

Given training set \(\mathcal{D} = \{z_i\}_{i \in |\mathcal{D}|}\), \(z_i\) for the fine-tuning stage can be further represented as \(x_i \cup y_i\) (\(x\), \(y\) respectively represent input and label). Both \(z_i\) and \(x_i\) can be denoted as a tokens sequence \(\{u_{j}\}_{j \in |x_i| or |z_i|}\). Generally following \cite{brown2020language}, the LM \(f_{\theta}\) is trained in two stages by maximizing the following likelihood separately:
\begin{equation}
\label{equation:likelihood}
    \begin{array}{c}
    L_{pt}(\mathcal{D}) = \sum_i \sum_j \log f_\theta(u_j|u_1,\cdots u_{j-1})\\
    L_{ft}(\mathcal{D}) = \sum_i \log f_\theta(y_i|u_1, \cdots, u_{|x_i|}) 
    \end{array}
\end{equation}
where $i, j$ respectively denote the index of the sample $z$ and the index of the token within $z$.

\textbf{Training Data Detection.} 
Following the settings of MIA \cite{yeom2018privacy, mattern2023membership}, for a given the target model $f_\theta$ and the sample $z$, the objective of training data detection can be defined as learning a detector \(\mathcal{A} : (z, f_\theta) \rightarrow \{0, 1\}\), where 1 denotes that $z$ is member data (\(z \in \mathcal{D}\)) and 0 denotes that \(z \notin \mathcal{D}\). As mentioned in Section ~\ref{section:rw_mia}, widely used calibration-free detection methods 
directly construct computational scores (denoted as $\mathcal{L}$), such as the loss, and threshold them:
\begin{equation}
    \begin{array}{c}
    \mathcal{A}(z, f_\theta) = \mathds{1}(\mathcal{L}(z,f_\theta)>\epsilon)
    \end{array}
\end{equation}

By comparing $\mathcal{L}$ with a predefined threshold $\epsilon$, $\mathcal{A}$ can achieve the detection of members or non-members. For more recent calibration-based methods, a calibration function \(c\) is additionally introduced to correct for the detector's bias relative to the target distribution. Calibration is typically achieved by correcting the original scores with the calibration function. The calibration function is usually a designed score \(c : (\tilde{z}, f_\phi) \rightarrow \mathbb{R}\) based on the differential performance on adjacent samples \(\tilde{z}\) \cite{mattern2023membership} or reference models \(f_\phi\) \cite{watson2021importance}. The reference models are usually additional models trained on a similar training data distribution. Consequently, the detector $\mathcal{A}$ can be extended as:
\begin{equation}
\label{equation:calibrated_mia}
    \begin{array}{c}
    \mathcal{A}(z, f_\theta) = \mathds{1}(\mathcal{L}(z,f_\theta)-c(\tilde{z}, f_\phi)>\epsilon)
    \end{array}
\end{equation}

Following the standard setup \cite{shi2023detecting}, we primarily constrain detecting pre-training samples under black-box settings. Specifically, the detection during the fine-tuning stage will be further discussed in Section ~\ref{subsection:2-stage}.

\subsection{Dynamic Benchmark Construction}
\label{subsection:benchmark}
StackMIA\footnote{The StackMIAsub benchmark dataset is available here: \href{https://huggingface.co/datasets/darklight03/StackMIA}{https://huggingface.co/datasets/darklight03/StackMIA}} is based on the Stack Exchange dataset \footnote{https://archive.org/details/stackexchange}, which is widely used for pre-training. Specifically, we organize member and non-member data with fine-grained release times to ensure reliability and applicability to newly released LLMs. 
More Details are provided in Appendix ~\ref{appendix:stackmia}.

\textbf{Data collection and organization.} We utilize the data source provided by the official (Appendix ~\ref{appendix:stackmia}). Each record contains a post and answers from different users. \textbf{Data collection:} Following the data selection strategy mentioned by LLMs such as LLaMA \cite{touvronllama}, etc, we retain data from the 20 largest websites, including common themes like English, math, etc. Subsequently, based on the training timelines of most LLMs \cite{zhao2023survey}, we set January 1, 2017, as the latest cutoff date for member data, i.e., data with both posts and answers dated before this time are considered members. For non-members, January 1, 2023, is set as the earliest occurrence time. Lastly, we build an automatic pipeline to remove HTML tags from the text. \textbf{Data filtering:} Considering the potential differences in answer sorting strategies \cite{ouyang2022training}, we only retain the post records. We apply automatic filtering based on: (1) posts contain only texts, not formulas or code; (2) posts have not been asked repeatedly. \textbf{Data organization:} To ensure applicability to a vast array of LLMs released after 2023, we reorganize the non-member data with fine-grained precision. We selected approximately 2000 posts created within each month, using the month as a cutoff point. This allows future users to construct subsets of StackMIA suitable for newly released LLMs through our provided pipeline.

\textbf{Benchmark test.}
Following the settings of \cite{shi2023detecting}, we divide the original WikiMIA by length. Simultaneously, we construct \textbf{StackMIAsub}\footnote{The StackMIAsub benchmark dataset is available here: \href{https://huggingface.co/datasets/darklight03/StackMIAsub}{https://huggingface.co/datasets/darklight03/StackMIAsub}} dataset (8267 samples in total, Appendix ~\ref{appendix:stackmia}) to conduct experiments in this paper. Specifically, we set May 1, 2023, as the cutoff date for non-member data. Following \cite{shi2023detecting}, we sequentially select approximately balanced sets of member and non-member data by length. Additionally, we use GPT-3.5-turbo\footnote{https://platform.openai.com/docs/api-reference} to construct (Appendix~\ref{appendix:stackmia}) synonymous rewritten data to test the stability of the methods under the condition of approximate memory \cite{ishihara2023training}. Referring to \cite{ippolito2022preventing}, we set BLEU \cite{papineni2002bleu}$>0.75$ as a condition to ensure semantic consistency in the rewritten data.

\section{Methodology}
\label{method}
We introduce \underline{\textbf{P}}olarized \underline{\textbf{A}}ugment \underline{\textbf{C}}alibration (\textbf{\ours}), an efficient and novel calibration-based training data detection method. 
The key idea is to construct adjacent samples using easy data augmentation for calibrating a generalized distribution and design a brand-new polarized distance to enhance the salience. We further propose a probabilistic tracking method suitable for models with partially inaccessible logits for the first time.

\subsection{Generating Adjacent Samples}
\label{subsection:adjacent}
As mentioned in Section ~\ref{subsection:formulation}, for a given data point $z$, we construct an adjacent sample space $\tilde{z}$ for the calibration function $c$. 
We opt for a simpler word-level perturbation approach through the Easy Data Augment \cite{wei2019eda} framework to generate $\tilde{z}$, which are adjacent in the local distribution with $z$. The calibration through these adjacent data points prevents calibration-free scores from being confused with misleading high-confidence, especially when the model provides a well-generalized distribution of non-members \cite{choquette2021label}. Specifically, we randomly swap 2 tokens from $z$, and repeat this process $m$ times:
\begin{equation}
\label{equation:calibrated_mia}
    \begin{array}{c}
    \tilde{z}=\sigma_m(z)= \sigma_m(\{u_{j}\}_{j\in [1, |z|]})
    \end{array}
\end{equation}
where $\sigma_m$ represents a bijection from z to itself (i.e., a permutation) after $m$ random swaps. 
Further, different from previous works, the augmentation-based scheme focuses the calibration on local distribution and is much more efficient due to the avoidance of introducing additional models.

\subsection{Polarized Distance}
\label{subsection:distance}
Due to the challenges in constructing reference models for LLMs, we calibrate directly using the difference between $\tilde{z}$ and $z$ without introducing costly external models. 
As mentioned in Section ~\ref{subsection:adjacent}, augmentation-based $\tilde{z}$ tends to exhibit non-member characteristics more. 
In practice, using traditional confidence scores, e.g. loss or perplexity, as the $\mathcal{L}$ score fails to demonstrate stable significance. \cite{shi2023detecting} proposes to improve classification effectiveness by calculating only a portion of the low-probability outlier words. We integrate this technique and expand it to focus simultaneously on far and close local regions of the token probability, achieving a significant measurement in probability space. 
Specifically, consider a sequence of tokens for a given sample $z$, denoted as $z = \{u_j\}_{j\in [1, |z|]}$. According to Equation ~\ref{equation:likelihood}, the log-probability of each token $u_i$ can be denoted as $\log f_\theta(u_i|u_1,\cdots,u_{i-1})$. As depicted in Figure ~\ref{fig:framework}, we then sort the probabilities of each token and select the largest $k_1\%$ and the smallest $k_2\%$ to form sets, denoted as $\mathrm{MAX}(z,k_1)$ and $\mathrm{MIN}(z, k_2)$, respectively.
Subsequently, the polarized distance $\mathcal{L}_M$ can be denoted as:
\begin{equation}
\label{equation:distance}
    \begin{array}{c}
    \mathcal{L}_M=\frac{1}{K_1}\sum\limits_{u_i\in \mathrm{MAX}(z,k_1)} \log f_\theta(u_i|u_1,\cdots,u_{i-1})\\
    -\frac{1}{K_2}\sum\limits_{u_i\in \mathrm{MIN}(z,k_2)} \log f_\theta(u_i|u_1,\cdots,u_{i-1})
    \end{array}
\end{equation}
where $K_1$ and $K_2$ denotes the size of $\mathrm{MAX}(z,k_1)$ and $\mathrm{MIN}(z,k_2)$ set separately.

\textbf{General.} According to the previous sections, the implementation of \ours{} can be represented as:
\begin{equation}
\label{equation:amd}
    \begin{array}{c}
    \mathcal{A}(z, f_\theta) = \mathds{1}[\mathcal{L}_M(z,f_\theta)-\sum\limits^{N}\frac{\mathcal{L}_M(\sigma_m(z), f_\phi)}{N}>\epsilon]
    \end{array}
\end{equation}
where $N$ denotes the number of repetitions to reduce random errors.

\subsection{Theoretical Analysis}
The explanation for the approach of \ours{} is quite straightforward: through carefully designed discrete perturbations, it makes $\tilde{z}$ (i.e., $\sigma_m(z)$) exhibit non-member characteristics. Since LLMs typically rely on original natural corpora, such perturbations can confuse the model \cite{jin2020bert, morris2020textattack, li2021contextualized}, thereby obtaining measurable differences between $z$ and $\tilde{z}$. By calculating the difference in the probability distribution of far and near local regions (expressed as the highest and lowest probabilities), $\mathcal{L}_M$ can reflect the model's predictive uncertainty \cite{duan2023shifting} and volatility. Given a sample $z$, when it occurs:
\begin{equation}
\label{equation:compare}
    \begin{array}{c}
    \mathcal{L}_M(z,f_\theta) \gg \mathcal{L}_M(\sigma_m(z),f_\theta)
    \end{array}
\end{equation}
This means that the impact of the perturbations is significant, which indicates that $z$ may be overfitting. In this case, $z$ will be classified as a member sample. Furthermore, we provide a simple detailed mathematical proof in Appendix ~\ref{appendix:proof}.

\subsection{Black-box Probabilistic Tracking}
Almost all detection methods require accessing the probabilities of all tokens for \(z\), which is not feasible for some current black-box models, such as GPT-4 \cite{achiam2023gpt} accessed by official API. These models only provide a log-probability query interface for the top \(n\) words, where \(n \leqslant 5\) usually. To address this issue, we take the GPT models as an example and construct a black-box probabilistic tracking algorithm using the \(\mathrm{logit\_bias}\)\footnote{https://platform.openai.com/docs/api-reference/completions/create\#completions-create-logit\_bias} function provided by the OpenAI API to track probability outputs. Such a function allows for setting biases of the logits for specific token IDs, which can be obtained through the tiktoken library\footnote{https://github.com/openai/tiktoken}. Utilizing this feature, for each token in turn \(u_i \in z\), we enumerate biases added to the corresponding token ID until the top \(n\) probability query results are just altered. The obtained bias threshold \(\gamma_i\) can be approximated as the difference between the log-probability of \(u_i\) and the known token \(u_\tau\) (see Appendix ~\ref{appendix:prob} for the proof). Thus, the probabilities of all tokens in \(z\) can be obtained by:
\begin{equation}
\label{equation:black_box}
    \begin{array}{c}
    \log f_\theta(u_i|\cdot) = \log f_\theta(u_\tau|\cdot)-\gamma_i
    \end{array}
\end{equation}
where $\cdot$ represents the prefix tokens of $u_i$. Due to the monotonicity of the impact of bias growth on query results, the enumeration process can be optimized for the binary search. Thus, we achieve a reduction in time complexity from $O(N)$ to $O(\log N)$ for obtaining the log probability of a single token with a logit of $N$, which is cost-effective and efficient. 

With this extraction method, \ours{} becomes the first method capable of detecting training data from almost any black-box LLMs.

\subsection{Pseudo Code}
\label{sec:pseudo_method}

We provide a simple pseudocode (as shown in Algorithm ~\ref{algorithm:pac}) to illustrate the specific implementation steps of \ours{}.

\begin{algorithm}
\caption{Polarized Augment Calibration}
\label{algorithm:pac}
\begin{algorithmic}[1]
\Statex \textbf{Input:} given data sample $z=\{u_i\}_{i\in|z|}$, source model $f_{\theta}$, and decision threshold $\epsilon$
\Statex \textbf{Output:} $\mathrm{True}$ – $z$ is a member sample, $\mathrm{False}$ – $z$ is a non-member sample.
\State Get the augmented sample $\tilde{z}$ with random swap, repeating $m$ times 
\State Select the highest $K_1$ probability tokens and lowerst $K_2$ probability tokens to construct $\mathrm{MAX}$ and $\mathrm{MIN}$ set
\State Calculate polarized distance $\mathcal{L}_M(z)$
\State $\mathcal{L}\gets\frac{1}{K_1}\sum\limits_{\mathrm{MAX}} \log f_\theta(u|\cdot)-\frac{1}{K_2}\sum\limits_{\mathrm{MIN}} \log f_\theta(u|\cdot)$
\State $\mathbf{d}\gets\mathcal{L}_M(z)-\mathcal{L}_M(\tilde{z})$ 
\State \textbf{return} $\mathrm{True}$ if $\mathbf{d}>\epsilon$ else $\mathrm{False}$
\end{algorithmic}
\end{algorithm}

\section{Experiments}
\subsection{Experiments Settings}
\label{subsection:setting}

\textbf{Baseline Methods}: We selected six popular methods to evaluate our approach: four calibration-based and two calibration-free. Calibration-based methods include: the Neighborhood attack (\textbf{Neighbor}) \cite{mattern2023membership}, which assesses loss differences between original samples and their neighbors generated by masked language models; and perplexity-based calibration \cite{carlini2021extracting} techniques utilizing Zlib entropy (\textbf{Zlib}) \cite{gailly2004zlib}, lowercased sample perplexity (\textbf{Lower}), and comparisons with reference models trained on the same dataset (\textbf{Ref}). Calibration-free methods comprise the \textbf{Min-K\%} method \cite{shi2023detecting}, predicting pre-trained samples through low-probability outlier words; and the Loss Attack \cite{yeom2018privacy}, substituting loss with Perplexity (\textbf{PPL}) in LLMs.

\textbf{Datasets and Metric.} We utilize the \textbf{StackMIAsub} benchmark (Section ~\ref{subsection:benchmark}) and the \textbf{WikiMIA} dataset proposed by \cite{shi2023detecting}.
WikiMIA (Appendix ~\ref{appendix:wikimia}) leverages Wikipedia timestamps and model release dates to identify member and non-member data sets, applicable for LLMs trained up to 2023. Both datasets are transformed into two formats as the guidelines in Section ~\ref{subsection:benchmark}: the original format (\textbf{ori}) and the synonym rewritten format (\textbf{syn}).

For evaluation, we follow \cite{mattern2023membership, carlini2022membership, watson2021importance} and plot the ROC curve analysis method. To facilitate numerical comparison, we primarily use the \textbf{AUC} score (Area Under the ROC Curve). The AUC score (Appendix ~\ref{appendix:auc}), independent of any specific threshold, accurately gauges the method's ability to differentiate between members and non-members. It also eliminates bias from threshold selection.

\textbf{Models.} We conduct experiments against 10 commonly used LLMs. Six models are applicable for both WikiMIA and StackMIA, including LLaMA-13B \cite{touvronllama}, LLaMA2-13B \cite{touvron2023llama}, GPT-J-6B \cite{gpt-j}, GPT-Neo-2.7B \cite{gpt-neo}, OPT-6.7B \cite{zhang2022opt}, and Pythia-6.9B \cite{biderman2023pythia}. The two GPT-3 base models, Davinci-002 and Baggage-002 \cite{ouyang2022training}, are suited for the WikiMIA dataset. Additionally, two newer models are applicable for StackMIA, including StableLM-7B \cite{stablelm} and Falcon-7B \cite{falcon40b}. 

\subsection{Implements} 
According to Section~\ref{method}, the key hyper-parameters affecting the \ours{} include the times of perturbations $m$, the tokens ratio in min-max distance $k_1$ and $k_2$, and the number of adjacent samples $N$. To ensure efficiency, $N$ is globally fixed at $5$. Based on this, we conduct a grid search \cite{liashchynskyi2019grid} on a reserved small-scale validation set of StackMIA using LLaMA-13B. The final settings are $k_1=5$, $k_2=30$, and $m=0.3 \times |z|$, where $|z|$ denotes the token number of $z$.

\subsection{\ours{} as A More Effective Detector}
Based on the settings described in Section~\ref{subsection:setting}, the primary comparison results between \ours{} and the baseline methods are listed in Table~\ref{table:main_result}. The experimental outcomes indicate that \ours{} consistently outperforms across all models and all data formats. Specifically, \ours{} shows an average AUC score improvement of \textbf{$4.5\%$} on WikiMIA and \textbf{$5.9\%$} on StackMIAsub compared to all other baseline methods. Moreover, \ours{} maintains robust performance even under the conditions of synonymously approximate memories. Notably, the Min-K\% method exhibits the second-best performance in all settings, validating the reliability of using local regions of token probabilities. And different from previous methods, \ours{} exhibits prominence in member recognition (Figure ~\ref{fig:intro_prob}). In summary, \ours{} is an effective and versatile solution for detecting pre-training data of LLMs.

\begin{table*}[htb]
  \centering
  \makebox[\linewidth]{
  \resizebox{\linewidth}{!}{%
  \begin{tabular}{lc|ccccccl|ccccccl}
    \toprule
    \multirow{2}{*}{Model} & \multirow{2}{*}{Form} & \multicolumn{7}{c|}{WikiMIA} & \multicolumn{7}{c}{StackMIAsub} \\
    \cline{3-9}\cline{10-16}
    & & PPL & Zlib & Lower & Ref & Neighbor & Min-K\% & \textbf{Ours} & PPL & Zlib & Lower & Ref & Neighbor & Min-K\% & \textbf{Ours} \\
    \midrule
    \midrule
    \multirow{2}{*}{LLaMA} & ori & 0.664 & 0.632 & 0.563 & 0.604 & 0.617 & 0.674 & $\textbf{0.706}^{\uparrow 4.7\%}$ & 0.605 & 0.556 & 0.532 & 0.487 & 0.552 & 0.612 & $\textbf{0.648}^{\uparrow 5.9\%}$  \\
    & syn & 0.684 & 0.627 & 0.545 & 0.587 & 0.610 & 0.681 & $\textbf{0.728}^{\uparrow 6.4\%}$ & 0.564 & 0.534 & 0.517 & 0.492 & 0.528 & 0.565 & $\textbf{0.592}^{\uparrow 4.7\%}$\\
    \midrule
    \multirow{2}{*}{LLaMA2} & ori & 0.540 & 0.555 & 0.520 & 0.540 & 0.509 & 0.535 & $\textbf{0.560}^{\uparrow 0.9\%}$ & 0.602 & 0.555 & 0.529 & 0.499 & 0.549 & 0.610 & $\textbf{0.643}^{\uparrow 5.4\%}$ \\
    & syn & 0.556 & 0.558 & 0.508 & 0.528 & 0.506 & 0.546 & $\textbf{0.572}^{\uparrow 2.5\%}$ & 0.563 & 0.533 & 0.519 & 0.507 & 0.525 & 0.565 & $\textbf{0.592}^{\uparrow 4.7\%}$\\
    \midrule
    \multirow{2}{*}{GPT-J} & ori & 0.641 & 0.620 & 0.558 & 0.616 & 0.631 & 0.675 & $\textbf{0.681}^{\uparrow 0.9\%}$ & 0.584 & 0.549 & 0.526 & 0.537 & 0.550 & 0.595 & $\textbf{0.605}^{\uparrow 1.7\%}$ \\
    & syn & 0.632 & 0.602 & 0.545 & 0.599 & 0.605 & 0.644 & $\textbf{0.666}^{\uparrow 3.4\%}$ & 0.544 & 0.525 & 0.507 & 0.548 & 0.518 & 0.549 & $\textbf{0.560}^{\uparrow 2.0\%}$\\
    \midrule
    \multirow{2}{*}{GPT-Neo} & ori & 0.616 & 0.603 & 0.560 & 0.593 & 0.619 & 0.648 & $\textbf{0.666}^{\uparrow 2.8\%}$ & 0.579 & 0.547 & 0.531 & 0.538 & 0.555 & 0.590 & $\textbf{0.600}^{\uparrow 1.7\%}$ \\
    & syn & 0.610 & 0.590 & 0.561 & 0.579 & 0.596 & 0.631 & $\textbf{0.653}^{\uparrow 3.5\%}$ & 0.539 & 0.523 & 0.507 & 0.545 & 0.520 & 0.545 & $\textbf{0.556}^{\uparrow 2.0\%}$\\
    \midrule
    \multirow{2}{*}{OPT} & ori & 0.602 & 0.591 & 0.560 & 0.633 & 0.577 & 0.625 & $\textbf{0.648}^{\uparrow 3.7\%}$ & 0.602 & 0.558 & 0.533 & 0.492 & 0.583 & 0.607 & $\textbf{0.619}^{\uparrow 2.0\%}$ \\
    & syn & 0.603 & 0.584 & 0.551 & 0.643 & 0.577 & 0.619 & $\textbf{0.646}^{\uparrow 4.4\%}$ & 0.559 & 0.534 & 0.518 & 0.508 & 0.545 & 0.560 & $\textbf{0.572}^{\uparrow 2.1\%}$ \\
    \midrule
    \multirow{2}{*}{Pythia} & ori & 0.635 & 0.617 & 0.550 & 0.629 & 0.626 & 0.664 & $\textbf{0.697}^{\uparrow 5.0\%}$ & 0.598 & 0.557 & 0.532 & 0.549 & 0.559 & 0.604 & $\textbf{0.624}^{\uparrow 3.3\%}$\\
    & syn & 0.634 & 0.602 & 0.549 & 0.623 & 0.614 & 0.642 & $\textbf{0.696}^{\uparrow 8.4\%}$ & 0.553 & 0.532 & 0.515 & 0.559 & 0.525 & 0.556 & $\textbf{0.578}^{\uparrow 4.0\%}$\\
    \midrule
    \multirow{2}{*}{StableLM} & ori & - & - & - & - & - & - & \multicolumn{1}{c|}{-} & 0.515 & 0.506 & 0.449 & 0.482 & 0.518 & 0.510 & $\textbf{0.589}^{\uparrow 14\%}$ \\
    & syn & - & - & - & - & - & - & \multicolumn{1}{c|}{-} & 0.491 & 0.488 & 0.437 & 0.484 & 0.501 & 0.487 & $\textbf{0.576}^{\uparrow 15\%}$\\
    \midrule
    \multirow{2}{*}{Falcon} & ori & - & - & - & - & - & - & \multicolumn{1}{c|}{-} & 0.613 & 0.566 & 0.519 & 0.577 & 0.573  & 0.617 & $\textbf{0.641}^{\uparrow 3.9\%}$ \\
    & syn & - & - & - & - & - & - & \multicolumn{1}{c|}{-} & 0.569 & 0.541 & 0.505 & 0.588 & 0.537 & 0.569 & $\textbf{0.593}^{\uparrow 0.8\%}$\\
    \midrule
    \multirow{2}{*}{davinci} & ori & 0.638 & 0.621 & 0.497 & 0.554 & 0.607 & 0.656 & $\textbf{0.694}^{\uparrow 5.8\%}$ & - & - & - & - & - & - & \multicolumn{1}{c}{-} \\
    & syn & 0.654 & 0.616 & 0.507 & 0.564 & 0.608 & 0.651 & $\textbf{0.691}^{\uparrow 5.6\%}$ & - & - & - & - & - & - & \multicolumn{1}{c}{-} \\
    \midrule
    \multirow{2}{*}{babbage} & ori & 0.569 & 0.575 & 0.492 & 0.475 & 0.537 & 0.559 & $\textbf{0.607}^{\uparrow 6.7\%}$ & - & - & - & - & - & - & \multicolumn{1}{c}{-} \\
    & syn & 0.582 & 0.576 & 0.513 & 0.483 & 0.540 & 0.574 & $\textbf{0.621}^{\uparrow 6.7\%}$ & - & - & - & - & - & - & \multicolumn{1}{c}{-} \\
    \midrule
    \multirow{2}{*}{Mean} & ori & 0.613 & 0.602 & 0.537 & 0.581 & 0.590 & 0.629 & $\textbf{0.657}^{\uparrow 4.5\%}$ & 0.587 & 0.549 & 0.519 & 0.520 & 0.554 & 0.593 & $\textbf{0.621}^{\uparrow 5.9\%}$ \\
    & syn & 0.619 & 0.594 & 0.535 & 0.576 & 0.582 & 0.623 & $\textbf{0.659}^{\uparrow 5.8\%}$ & 0.548 & 0.526 & 0.503 & 0.529 & 0.524 & 0.549 & $\textbf{0.577}^{\uparrow 5.1\%}$\\
    \bottomrule
  \end{tabular}%
  }
  }
  \caption{The AUC results of training data detection across various models on the WikiMIA and StackMIAsub. In particular, the percentage data represent the minimum percentage performance improvement of our \ours{} method.}
  \label{table:main_result}
\end{table*}

\subsection{Ablation Study}
\begin{table}[htb]
  \centering
  \resizebox{\linewidth}{!}{%
  \begin{tabular}{l|cccc|c}
    \toprule
    \multicolumn{5}{c}{Generation Method} \\
    \midrule
    Dataset & None & Neighbor & replace & delete & ours\\
    \midrule
    WikiMIA & -4.2\% & -7.5\% & -8.0\% & -7.2\% & 0.706 \\
    StackMIAsub & -5.4\% & -12.7\% & -6.9\% & -6.3\% & 0.648 \\
    \midrule
    \midrule
    \multicolumn{5}{c}{Metric Score} \\
    \midrule
    Dataset & PPL & Zlib & Min & Max & ours\\
    \midrule
    WikiMIA & -17.4\% & -17.4\% & -2.1\% & -14.4\% & 0.706 \\
    StackMIAsub & -22.3\% & -23.4\% & -0.7\% & -22.0\% & 0.648 \\
    \bottomrule
  \end{tabular}%
  }
  \caption{The AUC results on different generation methods and metric scores. `replace' and `delete' denote synonym replacement and random deletion, respectively. The scores not mentioned before include the log probability sum of tokens in high-probability regions.}
  \label{table:ablation}
\end{table}

To further validate the design of \emph{augment calibration} and \emph{polarized distance}, we conduct ablation studies on (1) methods for generating adjacent samples, and (2) metric scores. We limit the model to LLaMA-13B as an example.

\textbf{Generation method.} 
We evaluate the performance of four popular methods to generate adjacent samples. Table ~\ref{table:ablation} demonstrates a clear conclusion: augmentation based on random swaps is far more effective than any other generation method. We believe the underlying reason is that the swap operation ensures better non-member attributes, making the metric more significant.

\textbf{Metric Score.}
Similarly, we compared different metric scores in terms of their significance in difficulty calibration. As shown in Table~\ref{table:ablation}, the \emph{polarized distance} more readily facilitates the distinction between members and non-members.

\subsection{Analysis Study}
\begin{figure}[htb]
    \centering
    \includegraphics[width=\linewidth]{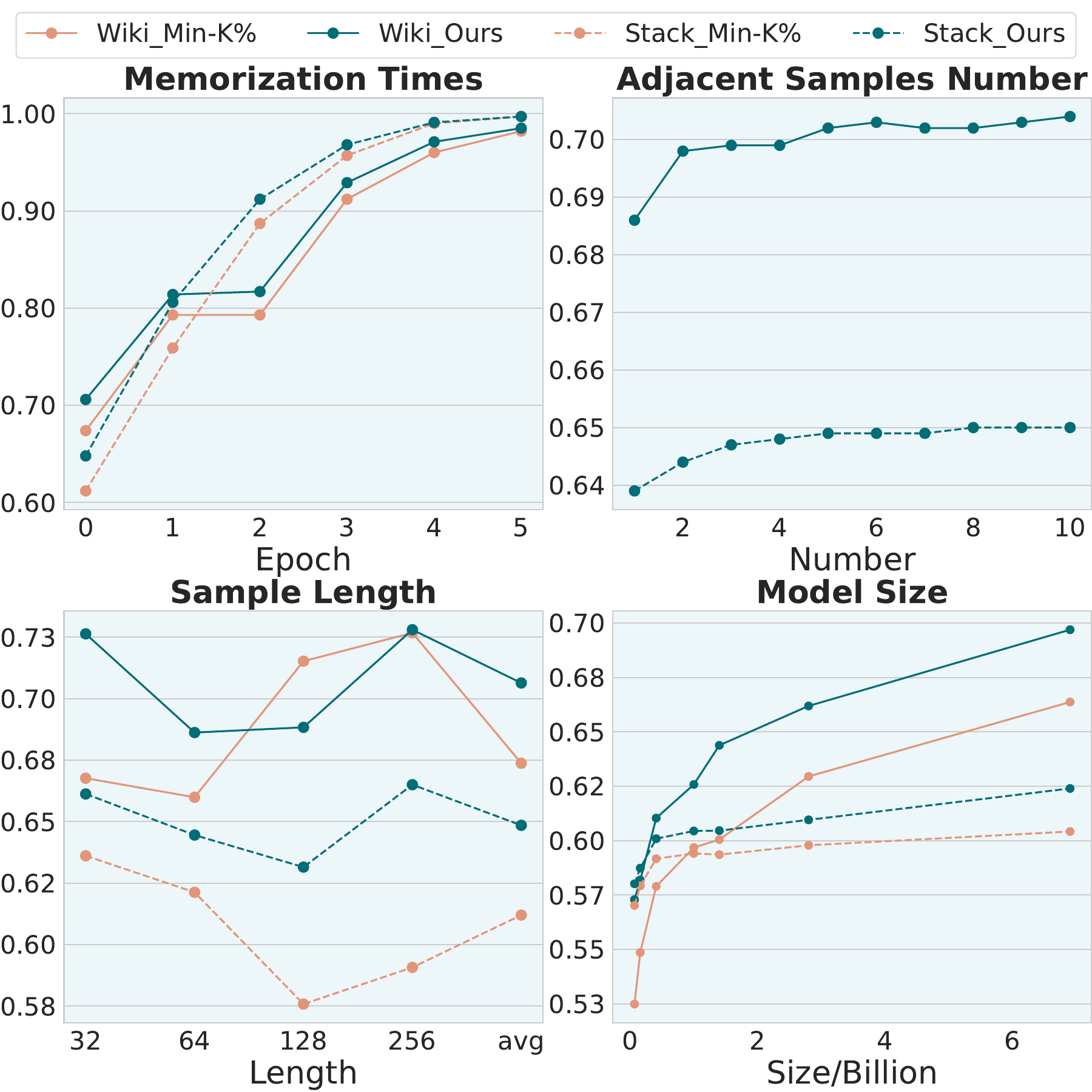}
    \caption{The AUC results as four different factors vary.}
    \label{fig:exp_analysis}
\end{figure}

To explore the factors influencing the detection, we focus on the four aspects, using LLaMA-13B and Pythia series as examples. All results are shown in Figure~\ref{fig:exp_analysis}.

\textbf{Memorization times.} 
We employ continual pre-training to increase the times that a member is seen by the model. The results show that detection difficulty decreases with memorization times increase, likely due to an increase in the degree of overfitting.

\textbf{Number of adjacent samples.} 
We varied the number of adjacent samples from one to twice the number of adjacent samples we used. The results demonstrate that with an increase in quantity, \ours{} maintain robust performance after initial improvements.

\textbf{Sample Length.} 
As the length of samples varies, \ours{} tends to achieve better performance in both lower and higher lengths. This is likely because they respectively contain more distinctive features and more information, making detection easier.

\textbf{Model Size.} 
The performance of \ours{} continuously improves with an increase in model parameters. This may be due to larger models having a stronger learning capacity within constant training iterations.

\subsection{\ours{} as A Two-Stage Detector}
\label{subsection:2-stage}
\begin{figure}[htb]
    \centering
    \includegraphics[width=\linewidth]{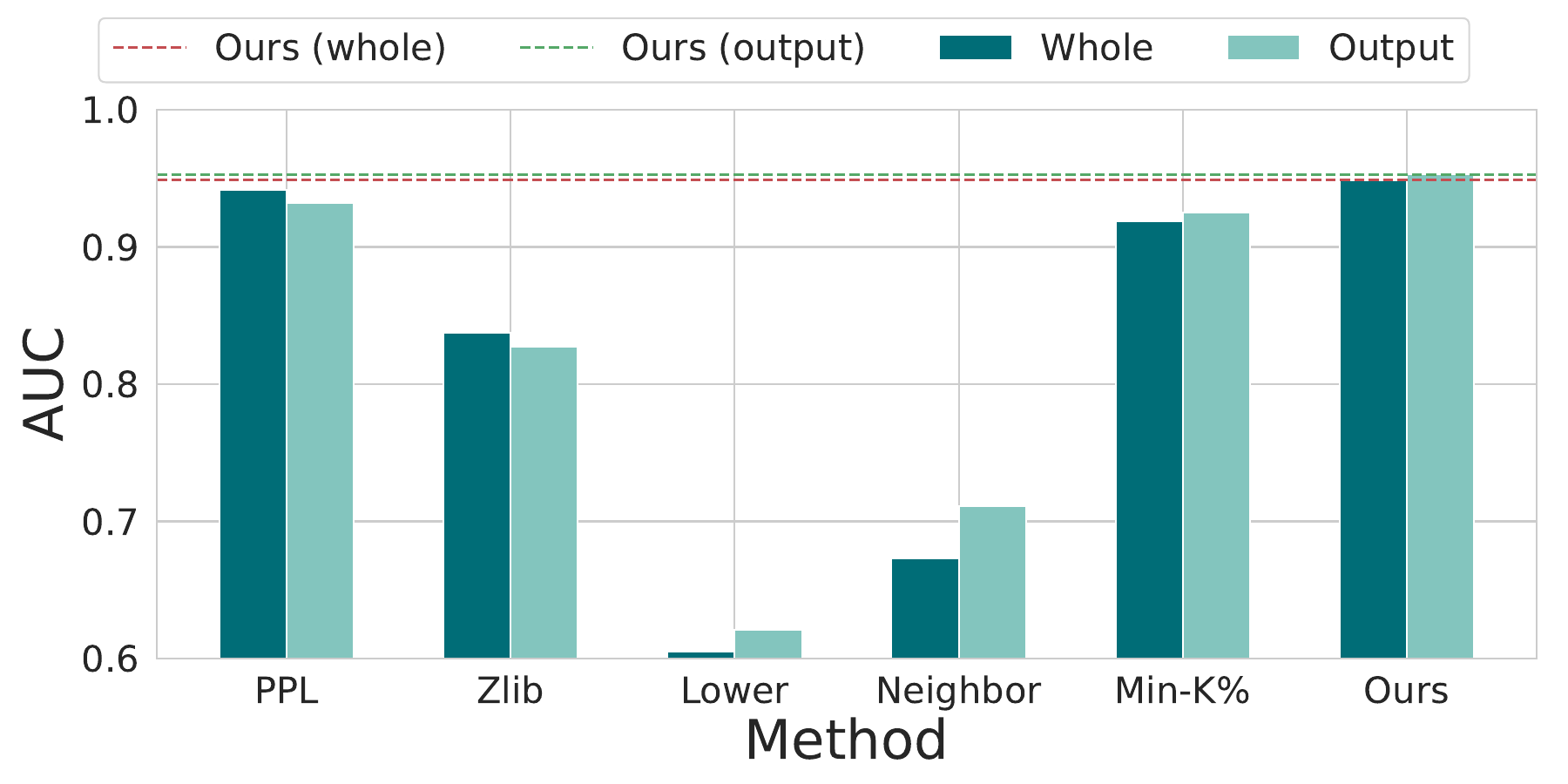}
    \caption{The AUC results in two-stage detection. `whole' and `output' represent two different settings of using the whole sample and the output part to detect.}
    \label{fig:2-stage}
\end{figure}

As more developers utilize various domain-specific data to fine-tune the same foundational models, the expansion of detection capabilities to the fine-tuning stage becomes increasingly critical. Limited works \cite{song2019auditing,mahloujifar2021membership} have addressed this issue but are not extendable to a two-stage process. We select a recent clear fine-tuning dataset after contamination check, Platypus \cite{platypus2023}, to fine-tune the LLaMA-13B model under a 5 epoch set. To further simulate real-world scenarios, we conducted detection with both output portions and entire samples. As Figure ~\ref{fig:2-stage} shows, \ours{} still exhibits excellent and stable performance even compared to the PPL score, which is directly equivalent to the training objective.

\subsection{\ours{} as A Robust Threshold Interpreter}
As mentioned in Section~\ref{subsection:formulation}, the selection of threshold $\epsilon$ affects the final detection effectiveness. Thus, we focus on the threshold obtained in the scenario where the knowledge about all samples is limited. We randomly select 10\%-50\% (in 5\% intervals) of the original dataset to form subsets. Then, following \citet{lipton2014thresholding}, we threshold \ours{} by maximizing the F1-score. As shown in Table~\ref{table:threshold}, \ours{} consistently outperforms baselines by at least \textbf{3\%} in accuracy. Simultaneously, the variance of the threshold $\epsilon$ obtained through subsets is relatively small (around 0.1), indicating that \ours{} requires only a small proportion of data to acquire a robust threshold, which is not limited by access-restricted data.

\begin{table}[htb]
  \centering
  \resizebox{\linewidth}{!}{%
  \begin{tabular}{lc|ccccc}
    \toprule
    Dataset & Metric & PPL & Zlib & Neighbor & Min-K\% & Ours\\
    \midrule
    \midrule
    \multirow{2}{*}{WikiMIA} & acc & 0.59 & 0.57 & 0.59 & 0.58 & \textbf{0.65}\\
     & std & 1.93 & \textbf{0.01} & 0.18 & 0.58 & $0.14^*$ \\
    \midrule
    \multirow{2}{*}{StackMIAsub} & acc & 0.54 & 0.52 & 0.52 & 0.55 & \textbf{0.58} \\
     & std & 1.32 & \textbf{0.03} & 0.18 & 0.25 & $0.09^*$ \\
     \midrule
    \multirow{2}{*}{Mix} & acc & 0.55 & 0.53 & 0.52 & 0.55 & \textbf{0.58} \\
     & std & 1.49 & \textbf{0.03} & 0.12 & 0.26 & $0.07^*$ \\
    \bottomrule
  \end{tabular}%
  }
  \caption{Results of threshold selection, where `acc' and `std' represent the accuracy and standard variance. The `Mix' denotes a mixed data set of the others.}
  \label{table:threshold}
\end{table}

\section{Case Study: Date Contamination}
\begin{figure}[htb]
    \centering
    \includegraphics[width=\linewidth]{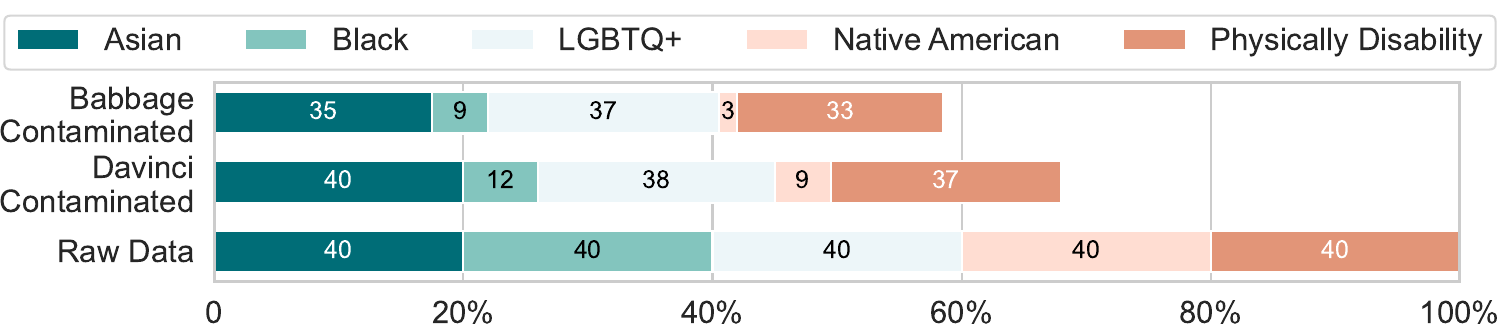}
    \caption{Bias data contamination cases of GPT-3 models. Cases are randomly selected from the TOXIGEN dataset.}
    \label{fig:case_toxigen}
\end{figure}

To further uncover the potential risks of existing LLMs (Large Language Models) through \ours{}, we selected two logical reasoning datasets, GSM8K \cite{cobbe2021training} and AQuA \cite{garcia2020dataset}, and one ethical bias investigation dataset, TOXIGEN \cite{hartvigsen2022toxigen} on GPTs LLMs. As depicted in Table~\ref{tab:case_study}, both GPT-3 and the more advanced ChatGPT and GPT-4 exhibited varying degrees of contamination, reaching up to \textbf{91.4\%} on davinci-002. Furthermore, all models unfortunately showed severe ethical bias data contamination in the training set (Figure ~\ref{fig:case_toxigen}). Based on the above, we call on the community to focus on finding solutions to the contamination problem to develop safer and more robust LLMs.

\begin{table}[htb]
  \centering
  \resizebox{\linewidth}{!}{%
  \begin{tabular}{l|cc|cc|cc|c}
    \toprule
    \multirow{2}{*}{Model} & \multicolumn{2}{c}{GSM8K} & \multicolumn{2}{c}{AQuA} & \multicolumn{2}{c|}{TOXIGEN} & \multirow{2}{*}{Avg} \\
    \cline{2-7}
    & Rate & Total & Rate & Total & Rate & Total & \\
    \midrule
    \midrule
    davinci-002 & 95.3\% & 1319 & 89.8\% & 254 & 45.5\% & 178 & 89.4\% \\
    \midrule
    babbage-002 & 84.6\% & 1319 & 72.4\% & 254 & 33.7\% & 178 & 77.7\% \\
    \midrule
    gpt-3.5-turbo & 82.0\% & 200 & 13.5\% & 200 & 5.06\% & 178 & 34.6\% \\
     \midrule
     GPT-4 & 64.0\% & 50 & 34.0\% & 50 & 6.7\% & 178 & 21.9\% \\
    \bottomrule
  \end{tabular}%
  }
  \caption{The cases of data contamination on GPTs. The results show both GPT-3 and the more advanced ChatGPT and GPT-4 exhibit varying degrees of contamination.}
  \label{tab:case_study}
\end{table}

\section{Conclusion}
We introduce Polarized Augment Calibration (\ours{}), a groundbreaking approach that expands the Membership Inference Attack (MIA) framework to detect training data in black-box LLMs. \ours{} unveils a new angle for MIA by utilizing confidence discrepancies across spatial data distributions and innovatively considering both distant and proximal probability regions to refine confidence metrics. This method is rigorously backed by theory and proven through comprehensive testing. We also present a novel detection technique for API-based black-box models using a proprietary probability tracking algorithm and launch StackMIA, a dataset aimed at overcoming the limitations of existing pre-trained data detection datasets. Applying \ours{} exposes widespread data contamination issues in even the most advanced LLMs, urging a communal effort towards addressing these challenges.

\section{Limitations}
While \ours{} shows promising results in detecting training data contamination in LLMs, its full potential is yet to be realized due to certain constraints. The limited availability of detailed training data information from LLMs providers restricts comprehensive validation across diverse models, underscoring the method's novelty yet implicating its untapped applicability. Additionally, the efficacy of \ours{} could be further enhanced with a more varied dataset, suggesting its adaptability and scope for refinement in varied LLMs. However, our current computational resources limit the extent of experiments, particularly on larger-scale LLMs, hinting at the method's scalability potential yet to be fully explored.

\bibliography{acl_natbib}
\newpage
\appendix

\section{Details of StackMIA}
\label{appendix:stackmia}
\subsection{Stack Exchange}

The Stack Exchange Data Dump contains user-contributed content on the Stack Exchange network. It is one of the largest publicly available repositories of question-answer pairs and covers a wide range of subjects --from programming to gardening, to Buddhism. Many large language models therefore include this dataset in their training data to enrich the training data and improve the model's ability to answer questions in different domains.

\subsection{StackMIAsub}

We follow \citet{shi2023detecting} to divide all the samples we collected into 4 categories based on length, ranging from 32 words to 256 words. The specific composition of StackMIA is shown in the Table ~\ref{appendix:Stackmia_composition}.
\begin{table}[htb]
  \centering
  \resizebox{\linewidth}{!}{%
  \begin{tabular}{l|ccccc}
    \toprule
    Length & 32 & 64 & 128 & 256 & Total \\
    \midrule
    \midrule
    Member & 1596 & 1740 & 680 & 90 & 4106 \\
    Non-member & 1615 & 1740 & 720 & 86 & 4161 \\
    \bottomrule
  \end{tabular}%
  }
  \caption{The samples composition of StackMIA dataset.}
  \label{appendix:Stackmia_composition}
\end{table}

\subsection{Synonymous Rewritten Data}
As mentioned in Section ~\ref{subsection:benchmark}, we rewrite the original samples to simulate approximate memorization scenarios by prompting GPT-3.5-turbo API.
Part of the prompts we used to rewrite the sentence are listed as follows:

\begin{itemize}
    \item ``Rewrite the sentence with the smallest possible margin, keeping the same semantics and do not complete anything, ensuring that BLEU > 0.75 before and after the change: `\textit{text}' \textquotedblright{}
    \item ``Slightly rewrite the following sentence without changing the sentence structure, and do not complete any sentence: `\textit{text}' \textquotedblright{}
    \item ``Randomly replace 2 words in the giving sentence with synonyms: `\textit{text}' \textquotedblright{}
    \item ``Replace 5 percent of the prepositions with a synonym in the giving sentence: `\textit{text}' \textquotedblright{}
\end{itemize}

The case comparison of the samples before and after the synonymous transformation with ChatGPT is listed in Table ~\ref{appendix:rewritten_prompt}.

\begin{table}[htb]
  \centering
  \resizebox{\linewidth}{!}{%
  \begin{tabular}{l|ccccc}
    \toprule
    Length & 32 & 64 & 128 & 256 & Total \\
    \midrule
    \midrule
    Member & 387 & 284 & 139 & 51 & 861 \\
    Non-member & 389 & 258 & 111 & 31 & 789 \\
    \bottomrule
  \end{tabular}%
  }
  \caption{The samples composition of WikiMIA Dataset.}
  \label{appendix:wikimia_composition}
\end{table}

\begin{table*}[htb]
  \centering
  \resizebox{\linewidth}{!}{%
  \begin{tabular}{p{0.1\textwidth}|p{0.9\textwidth}}
    \toprule
    ori & Density Functional Theory (DFT) is formulated to obtain ground state properties of atoms, molecules and condensed matter. However, why is DFT not able to predict the exact band gaps of semiconductors and insulators? Does it mean that the band gaps of semiconductors and insulators are not the ground states? \\
    \midrule
    syn & Why is it that Density Functional Theory (DFT) cannot accurately predict the precise band gaps of semiconductors and insulators, even though it is designed to determine the ground state properties of atoms, molecules, and condensed matter? Does this imply that the band gaps of semiconductors and insulators are not considered as ground states? \\
    \midrule
    \midrule
    ori & I am currently studying Electrical \& Electronic Engineering. I wish to pursue Quantum Mechanics or Quantum Computing as my research subject. Is it possible for me to do my M.Tech. and then pursue my research subject? What are the prerequisites for studying these subjects? I would be grateful if you could help me. \\
    \midrule
    syn & I am presently studying Electrical \& Electronic Engineering. I desire to pursue Quantum Mechanics or Quantum Computing as my research topic. Is it feasible for me to do my M.Tech. and then pursue my research topic? What are the requirements for studying these subjects? I would be thankful if you could assist me. \\
    \midrule
    \midrule
    ori & How is the meaning of a sentences affected by chosing one of those words? For instance, what's the different between The screech cicadas reverberated through the forest. and The screech cicadas reverberated throughout the forest. \\
    \midrule
    syn & How does the choice of one of those words affect the meaning of a sentence? For example, what is the difference between \"The screech cicadas reverberated through the forest.\" and \"The screech cicadas reverberated throughout the forest.\"? \\
    \midrule
    \midrule
    ori & The majority of definitions give the same meaning - \"Pandora's box\" is a synonym for \"a source of extensive but unforeseen troubles or problems.\" Are there any other metaphors or phrases with the same meaning? \\
    \midrule
    syn & Do any other metaphors or phrases convey the same meaning as the majority of definitions, which state that \"Pandora's box\" is synonymous with \"a source of extensive but unforeseen troubles or problems\"? \\
    \midrule
    \midrule
    ori & The majority of definitions give the same meaning - \"Pandora's box\" is a synonym for \"a source of extensive but unforeseen troubles or problems.\" Are there any other metaphors or phrases with the same meaning? \\
    \midrule
    syn & Do any other metaphors or phrases convey the same meaning as the majority of definitions, which state that \"Pandora's box\" is synonymous with \"a source of extensive but unforeseen troubles or problems\"?" \\
    \bottomrule
  \end{tabular}%
  }
  \caption{The cases before and after are synonymous rewritten with ChatGPT. The listed cases are selected from the member data with a length between 32 and 64.}
  \label{appendix:rewritten_prompt}
\end{table*}

\section{Details of WikiMIA}
\label{appendix:wikimia}
WikiMIA is a benchmark for MIA(\cite{shi2023detecting}), with data sourced from Wikipedia. For non-member data, the dataset collects recent event pages using January 1, 2023, as the cutoff date. For member data, the dataset collects articles before 2017. The specific composition of WikiMIA is shown in the Table ~\ref{appendix:wikimia_composition}.

\section{A Simple Proof of PAC method}
\label{appendix:proof}
Here we provide a mathematical proof why our method can effectively distinguishes members and non-members. 

As mentioned in Equation ~\ref{equation:distance}, given a sample $z$, the Polarized Distance can be briefly represented as the difference between the two terms $\mathrm{T}_1$ and $\mathrm{T}_2$:
\begin{equation}
    \begin{array}{c}
    \mathrm{T}_1 = \frac{1}{K_1}\sum\limits_{u_i\in \mathrm{MAX}(z,k_1)} log f_\theta(u_i|u_1,\cdots,u_{i-1})\\
    \mathrm{T}_2 = \frac{1}{K_2}\sum\limits_{u_i\in \mathrm{MIN}(z,k_2)} log f_\theta(u_i|u_1,\cdots,u_{i-1})
    \end{array}
\end{equation}
Each term can be scaled to the following inequality:
\begin{equation}
    \begin{array}{c}
    \frac{|z|}{K_1}\mathrm{T}_1 \geqslant \mathbb{E}_{u\in z}[\log f_\theta(u|\cdot)]\\
    \frac{|z|}{K_2}\mathrm{T}_2 \leqslant \mathbb{E}_{u\in z}[\log f_\theta(u|\cdot)]
    \end{array}
\end{equation}
where $\mathbb{E}$ function denotes the expectation function and $\cdot$ denotes the prefix tokens sequence. Then the MinMax Distance can be represented as:
\begin{equation}
    \begin{array}{c}
    0 \geqslant \mathcal{L}_M=\mathrm{T}_1-\mathrm{T}_2 \geqslant \frac{K_1-K_2}{|z|}\mathbb{E}_{u\in z}[\log f_\theta(u|\cdot)]
    \end{array}
\end{equation}
Then Equation ~\ref{equation:amd} can be further converted to:
\begin{equation}
    \begin{array}{c}
    \frac{K_1-K_2}{|z|}\mathbb{E}_{u\in z}[\log f_\theta(u|\cdot)] \leqslant\\
    \mathcal{L}_M(z)-\mathcal{L}_M(\sigma_m(z)) \leqslant \\
    \frac{K_2-K_1}{|\sigma_m(z)|}\mathbb{E}_{u\in \sigma_m(z)}[\log f_\theta(u|\cdot)]
    \end{array}
\end{equation}
Since the coefficient is constant, the threshold interval of the final calculated indicator can be equivalently expressed as:
\begin{equation}
\label{equation:proof_full_term}
    \begin{array}{c}
    \mathbb{E}_{u\in z}[\log f_\theta(u|\cdot)]+\mathbb{E}_{u\in \sigma_m(z)}[\log f_\theta(u|\cdot)]
    \end{array}
\end{equation}
Among them, the former term can be considered as a variant of the LLM training objective. Therefore, the value of Formula ~\ref{equation:proof_full_term} will change significantly as $z$ is fitted as member data. Moreover, there exists a positive correlation between the posterior term and the forward tendency according to \citet{jin2020bert}. This means that the threshold intervals of members and non-members are more sparse than solely using the probability of $z$ directly. Therefore, our indicator significantly captures differences between members and non-members.

\section{Proof of Probability Extraction}
\label{appendix:prob}
Assume that the logit output of the model $f_\theta(u_i|\cdot)$ is $l_1,\cdots,l_N$, and then the log-probability of $u_i$ can be represented as:
\begin{equation}
    \begin{array}{c}
    \log f_\theta(u_i|\cdot)=\log \frac{e^{l_i}}{\sum_{j=1}^{M} e^{l_j}}=l_i - \log\sum_{j=1}^{N} e^{l_j}
    \end{array}
\end{equation}
Then, the log-probability $\log f'_\theta(u_i|\cdot)$ after adding a fixed bias $\gamma_i$ to the logit of $u_i$ can be calculated as:
\begin{equation}
    \begin{array}{c}
    \log f'_\theta(u_i|\cdot)\approx\log \frac{e^{l_i + \gamma_i}}{\sum_{j=1}^{N} e^{l_j}} \\
    = (l_i + \gamma_i) - \log\sum_{j=1}^{N} e^{l_j}
    \end{array}
\end{equation}
Then the original log-probability can be calculated as:
\begin{equation}
    \begin{array}{c}
    \log f_\theta(u_i|\cdot)\approx\log f'_\theta(u_i|\cdot)-\gamma_i
    \end{array}
\end{equation}

\section{Baseline Details}
\label{appendix:baseline}
\subsection{Reference model comparison}
Reference model-based methods target at training reference models in the same manner as the target model (e.g., on the shadow data sampled from the same underlying pretraining data distribution). The raw score of the original samples can be calibrated with the average of the score in these reference models. Due to the high cost of strictly training a shadow LLM, we follow \citet{carlini2021extracting} to choose a much smaller model trained on the same underlying dataset. Specifically, we choose LLaMA-7B as the reference model of LLaMA-13B, LLaMA2-7B for LLaMA2-13B, GPT-Neo-125M for GPT-Neo-2.7B, OPT-125M for OPT-6.7B, Pythia-70M for Pythia-6.9B, StableLM 3B for StableLM-7B. Specially, for those LLMs without smaller models in the series, we use an approximate model trained on the mentioned same-dataset or distribution in their official statement, including GPT-Neo-125M for GPT-J-6B both trained on the Pile \cite{gaopile}, GPT-Neo-125M for Falcon-7B trained on the Pile and GPT2-124M \cite{radford2019language} for two OpenAI base models (Davinci-002 and Babbage-002).

\subsection{Neighborhood Attack via Neighbourhood Comparison}

This method is proposed by \citet{mattern2023membership}. The main idea is to compare the neighbors' losses and those of the original sample under the target model by computing their differences.

In our replication process, we followed the method on official Github \footnote{https://github.com/mireshghallah/neighborhood-curvature-mia} to select key hyper-parameters. Specifically, We randomly mask 30 percent of the tokens in the original sample with a span length of 2 to generate 25 masked sentences from the original sample and then use the T5-3b mask-filling model to generate 25 neighbors.

To evaluate the score of a sample, we use the formula as follows:
\[
\frac{\mathcal{L}(x) - \sum\limits^N \frac{\mathcal{L}(\tilde{x})}{N}}{\sigma}
\]
where $\sigma$ is the standard deviation of the neighbours' losses.

\subsection{Min-K\%}
The Min-K\% method, proposed by \citet{shi2023detecting}, is quite straightforward. It builds on the hypothesis that a non-member example is more likely to include a few outlier words with high negative log-likelihood (or low probability), while a member example is less likely to include words with high negative log-likelihood.
Specifically, Min-K\% is calculated as :
\begin{equation}
\mathrm{model}(x)= \frac{1}{E} \sum_{x_i \in \mathrm{Min-K\%}(x)}{ \log p(x_i | x_1, ..., x_{i-1})}.
\end{equation}
where $x = x_1, x_2, ..., x_N$ is the tokens in the sentences, while $\log p(x_i | x_1, ..., x_{i-1})$ is the log-likelihood of a token, $x_i$. 

In particular, we follow the original paper to set $K = 20$ for detection in experiments.

\section{Descriptions of AUC score}
\label{appendix:auc}
To evaluate with AUC score, we first plot the ROC curve through the True Positive Rate (TPR) and False Positive Rate(FPR). The ROC curve is used to plot TPR versus FPR using different classification thresholds. Lowering the classification threshold causes more categories of items to be classified as positive, thus increasing the number of false positives and true examples. AUC is then defined as the Area Under the ROC curve, providing an aggregate measure of the effect of all possible classification thresholds.

\section{Sentiment Analysis}

\begin{figure}[htb]
    \centering
    \includegraphics[width=\linewidth]{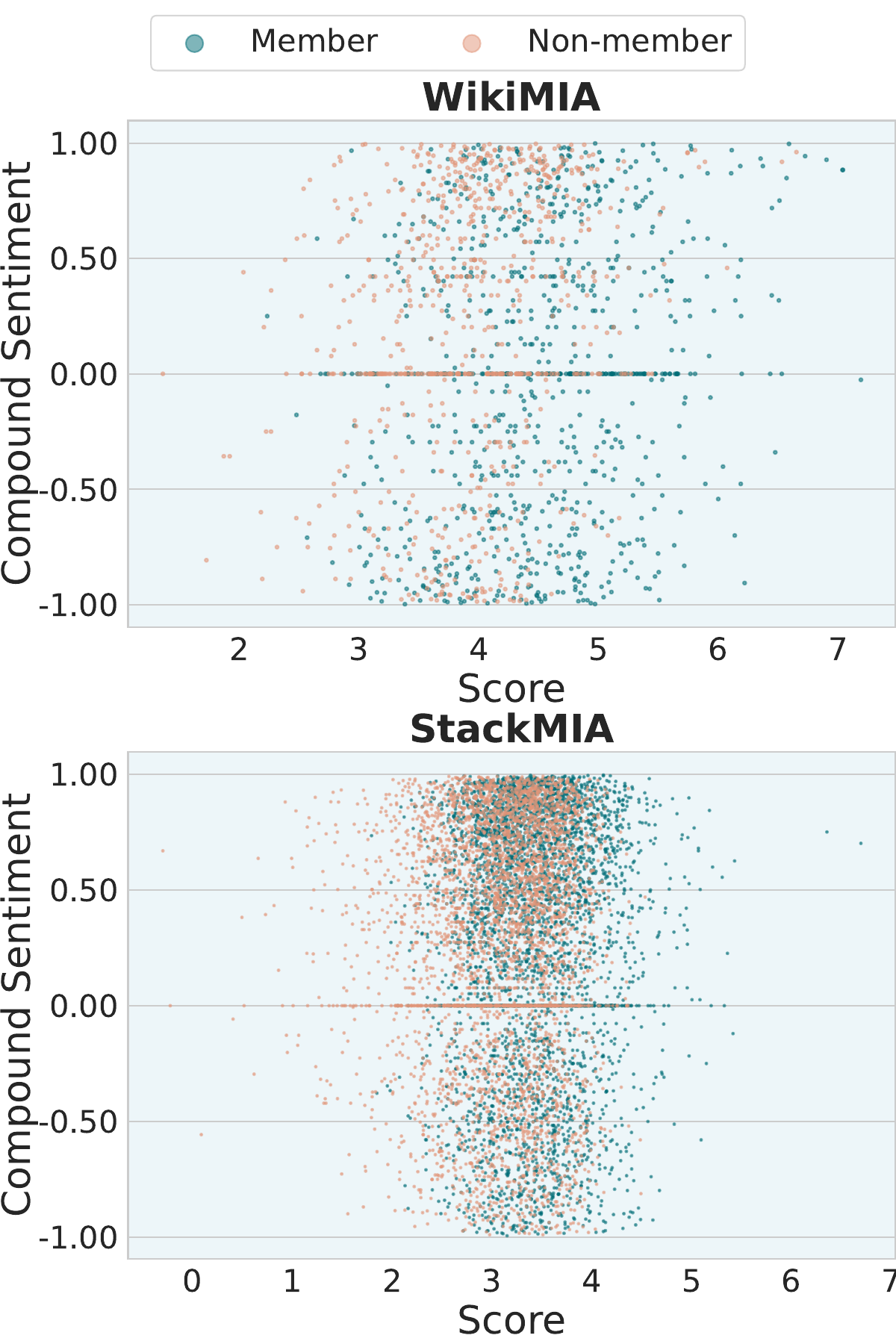}
    \caption{Sentiment analysis on WikiMIA \& StackMIA datasets. The X-axis represents the prediction score given by AMD, and the Y-axis represents the sentiment analysis score, with higher scores meaning positive and lower scores meaning negative.}
    \label{fig:sentiment-scatter}
\end{figure}

We further conducted syntax analysis experiments on both member and non-member samples. The specific results are shown in Figure ~\ref{fig:sentiment-scatter}.

\section{More Experiments Demonstrations}

\begin{figure*}
    \centering
    \begin{minipage}[!t]{0.49\linewidth}
        \centering
        \includegraphics[width=\linewidth]{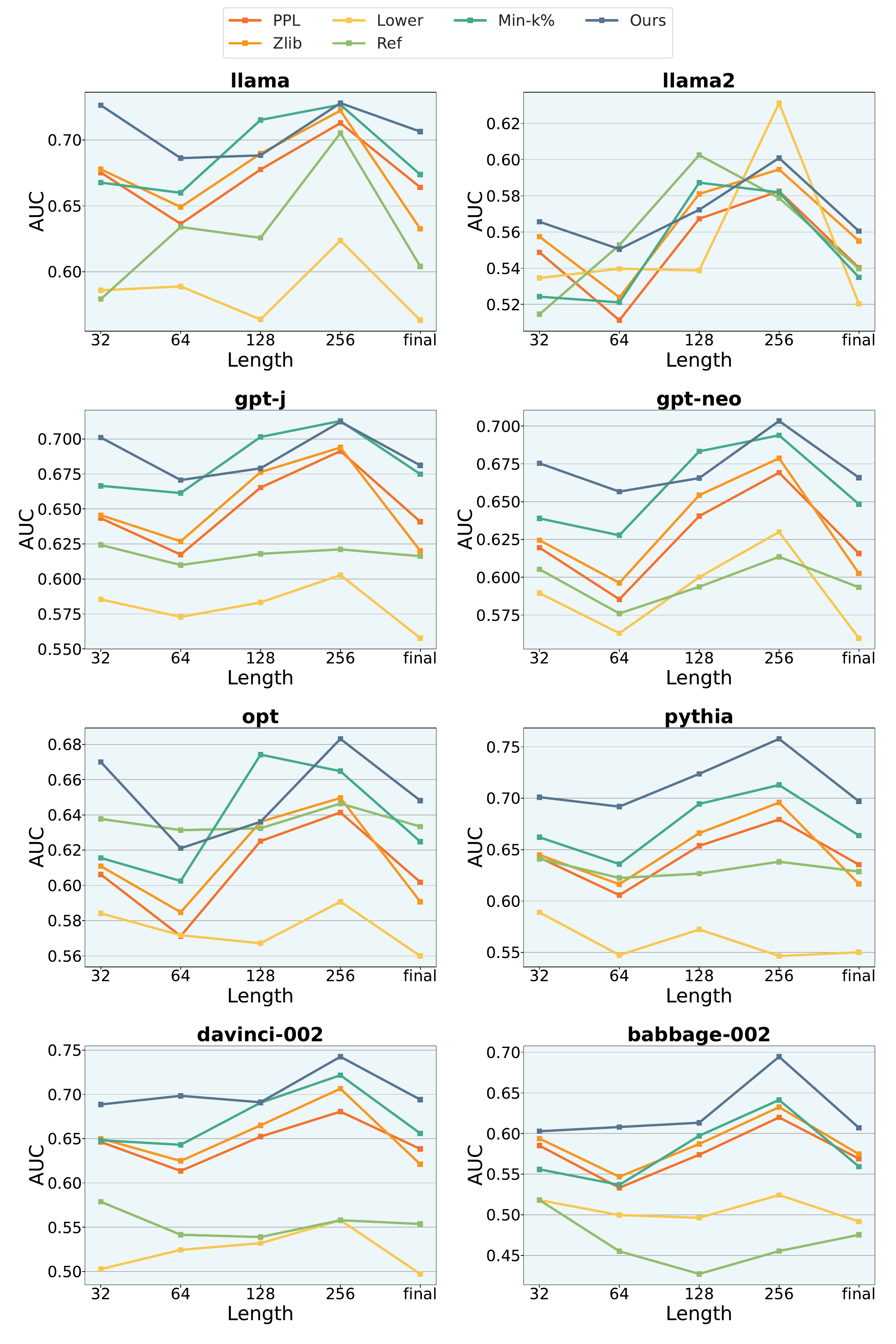}
        \caption{The AUC results of training data detection across various models with different methods on the WikiMIA dataset.}
        \label{fig:wikimia_merge}
    \end{minipage}
    \hfill
    \begin{minipage}[!t]{0.49\linewidth}
        \centering
        \includegraphics[width=\linewidth]{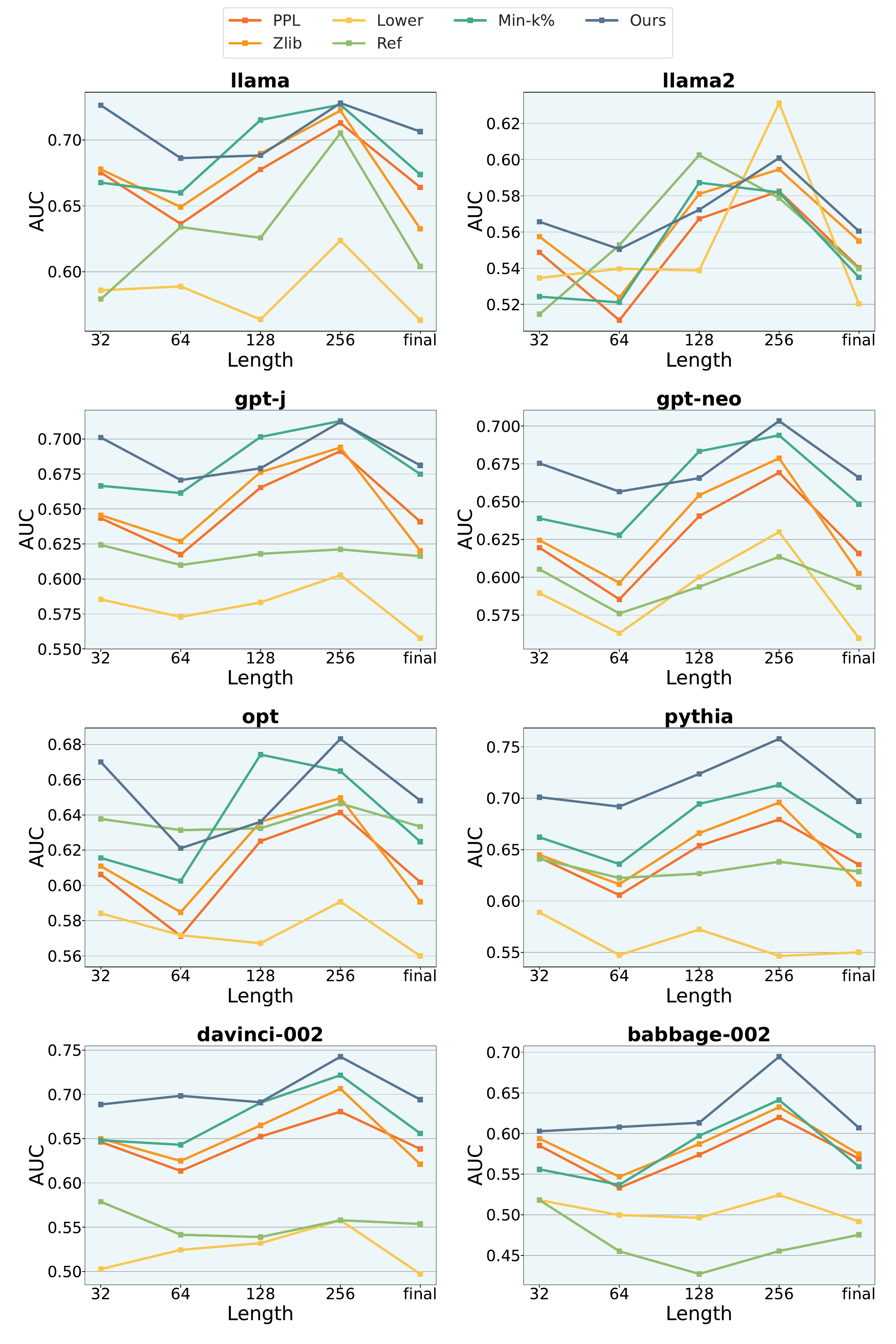}
        \caption{The AUC results of training data detection across various models with different methods on the StackMIAsub dataset.}
        \label{fig:stack_merge}
    \end{minipage}
\end{figure*}

The visualizing AUC results of training data detection across various models with different methods are shown in Figure ~\ref{fig:wikimia_merge} \& ~\ref{fig:stack_merge}.

\end{document}